%% file: Article_EndToEndP300BCI_arxiv.tex
\documentclass[journal]{IEEEtran} 
\usepackage{cite}
\ifCLASSINFOpdf
   \usepackage[pdftex]{graphicx}
	\DeclareGraphicsExtensions{.eps,.png}
\else
  \usepackage[dvips]{graphicx}
  \usepackage{pgf}
\fi
\usepackage[cmex10]{amsmath}
\usepackage{array}
\usepackage[font=footnotesize]{caption}
\ifCLASSOPTIONcompsoc
  \usepackage[caption=false,font=normalsize,labelfont=sf,textfont=sf]{subfig}
\else
  \usepackage[caption=false,font=footnotesize]{subfig}
\fi
\usepackage{pgf}
\usepackage{amssymb,amsfonts}
\usepackage{multirow}
\usepackage{color}
\usepackage{algorithm}
\usepackage{algpseudocode}
\usepackage{siunitx} 
\usepackage{paralist}
\usepackage{booktabs}
\usepackage{tabularx}
\usepackage{url}
\usepackage[colorlinks=true,citecolor=blue,urlcolor=blue,linkcolor=blue]{hyperref}
\usepackage{bbm}

\usepackage{lineno}
\usepackage{xcolor}
\newcommand\hl[1]{#1}%

\newcommand{\Ma}{\ensuremath{\mathcal{M}}} 

\newcommand{\kt}{\ensuremath{\text{T}}}
\newcommand{\knt}{\ensuremath{\text{NT}}}

\newcommand{\OurModel}{ASAP}
\newcommand{\MdmModel}{MDM+OM}
\newcommand{\XdawnModel}{xDAWN+OM}
\newcommand{\LdaModel}{RegLDA+OM}

\begin{document}
\title{End-to-end P300 BCI using Bayesian accumulation\\of Riemannian probabilities}

\author{Quentin~Barth\'elemy, Sylvain~Chevallier, Rapha\"elle~Bertrand-Lalo, Pierre Clisson

\thanks{Q. Barth\'elemy (ORCiD: 0000-0002-7059-6028) (corresponding author) is with Foxstream,
6 rue du Dauphin\'e, 69120 Vaulx-en-Velin, France (e-mail: q.barthelemy@gmail.com).}
\thanks{S. Chevallier (ORCiD: 0000-0003-3027-8241) is with Universit\'e Paris-Saclay,
UVSQ, LISV, 78124, V\'elizy-Villacoublay, France (e-mail: sylvain.chevallier@uvsq.fr).}
\thanks{R. Bertrand-Lalo is an independent scientist, Nantes, France (r.bertrand.lalo@gmail.com).}
\thanks{P. Clisson (ORCiD: 0000-0002-3123-9597) is an independent scientist, Paris, France (pierre@clisson.com).}
}

\maketitle

\begin{abstract}
In brain-computer interfaces (BCI), most of the approaches based on event-related potential (ERP) focus on the detection of P300,
aiming for single trial classification for a speller task.
While this is an important objective, existing P300 BCI still require several repetitions to achieve a correct classification accuracy.
Signal processing and machine learning advances in P300 BCI mostly revolve around the P300 detection part,
leaving the character classification out of the scope.
To reduce the number of repetitions while maintaining a good character classification,
it is critical to embrace the full classification problem.
We introduce an end-to-end pipeline, starting from feature extraction, and composed of
an ERP-level classification using probabilistic Riemannian MDM
which feeds a character-level classification using Bayesian accumulation of confidence across trials.
Whereas existing approaches only increase the confidence of a character when it is flashed,
our new pipeline, called B\underline{a}ye\underline{s}ian \underline{a}ccumulation of Riemannian \underline{p}robabilities (\OurModel), update the confidence of each character after each flash.
We provide the proper derivation and theoretical reformulation of this Bayesian approach for a seamless
processing of information from signal to BCI characters.
We demonstrate that our approach performs significantly better than standard methods on public P300 datasets.
\end{abstract}


\IEEEpeerreviewmaketitle


\section{Introduction}

Brain-computer interfaces (BCI) have become a rising domain in neurotechnologies, allowing decoding brain activity,
with multiple applications~\cite{Nam2018book} using electroencephalography (EEG).
Out-of-the-lab applications in EEG-based BCI face a difficult challenge:
on top of the hard problem of decoding brain signals in real-time,
most of the existing literature provides results that are difficult or even impossible to reproduce with online applications.
A large portion of the BCI literature tackles processing or classification problem relying on offline analysis and dataset-specific preprocessing,
or uses machine learning methods not suitable for real-world applications (long calibration process and model training phase).
Indeed, the BCI community is continuously pushing toward increased reproducibility, robustness and acceptability of BCI,
but the above-mentioned pitfalls are well documented~\cite{Huggins2014,Brunner2015,congedo2017riemannian,Rashid2020}.

In this study, we focus on a well-known class of BCI,
that is based on the detection of event-related potentials (ERP) induced by an oddball effect.
Common applications are P300-spellers~\cite{Farwell1988,Rivet2009,barachant2014plug} or P300-games,
such as Brain Invaders~\cite{barachant2014plug} or Raccoons vs Demons~\cite{Goncharenko2020Raccoons}.
P300 is an ERP elicited when the BCI flashes the target character (also called symbol or item).
The BCI's task in these ERP-based paradigms is to detect the presence or absence of a P300
to retrieve the flashed character~\cite{Farwell1988}.

The low signal-to-noise ratio (SNR) of P300 waves in EEG signals makes it difficult to classify them on a single trial.
Therefore, the first attempts to detect P300 waves consisted in averaging multiple trials (about 15 repetitions for each flashing character).
While these attempts were able to successfully discriminate between target and non-target flashing~\cite{Farwell1988},
the resulting BCI was slow and the information transfer rate (ITR) \cite{Yuan2013} was low.
To reduce the number of repetitions, further attempts applied advanced spatial filters to increase the amplitude related to P300 waves,
like xDAWN which maximizes the signal to signal+noise ratio~\cite{Rivet2009}.
To further reduce the required number of repetitions, ERP detection was achieved with classifiers such as LDA or SVM trained on signals enhanced with spatial filters~\cite{Lotte2018}.

However, spatial filters are subject- and session-dependent~\cite{Tomioka2007,Reuderink2008,Grosse2009},
preventing cross-session and cross-subject transfer capabilities~\cite{barachant2014plug}.
This implies that a subject should perform a careful calibration phase before being able to use the BCI;
this calibration is required for each session even for someone using a BCI on a regular basis.
Recently, Riemannian approaches applied to BCI have allowed the design of new classifiers,
such as the minimum-distance-to-mean (MDM)~\cite{barachant2014plug},
having high performances with shorter calibrations~\cite{barachant2014plug,congedo2017riemannian}.
Whereas optimized spatial filters estimate enhanced signals,
Riemannian BCI work in the space of covariance matrices to process and classify the signal.
While covariance captures the amplitude variations well, as induced by motor imagery tasks~\cite{Barachant2012},
it does not apply straightaway for ERP.
It is possible to extend the covariance matrix with a P300 prototype~\cite{barachant2014plug}
to obtain a discriminative representation that is appropriate for ERP classification.
When applying those prototype-based covariance matrices,
Riemannian BCI increased the robustness of P300-based BCI~\cite{barachant2014plug,korczowski2015single,congedo2017riemannian}.
It should be noted that these works focus on a signal level, \textit{i.e.} ERP detection,
and not at a BCI system level, for character prediction.

%

In P300 BCI, the aim is to predict character from ERP classification.
The most common approach is to increase the confidence of flashed characters if a P300 is detected.
Since each character is seldom flashed, as expected in an oddball paradigm,
and multiple repetitions are necessary to attain a reliable confidence in classification,
the transformation from ERP detection to character prediction is a bottleneck that hinders the BCI speed.
%

\begin{figure}[t]
	\centering
	\pgfimage[width=0.95\linewidth]{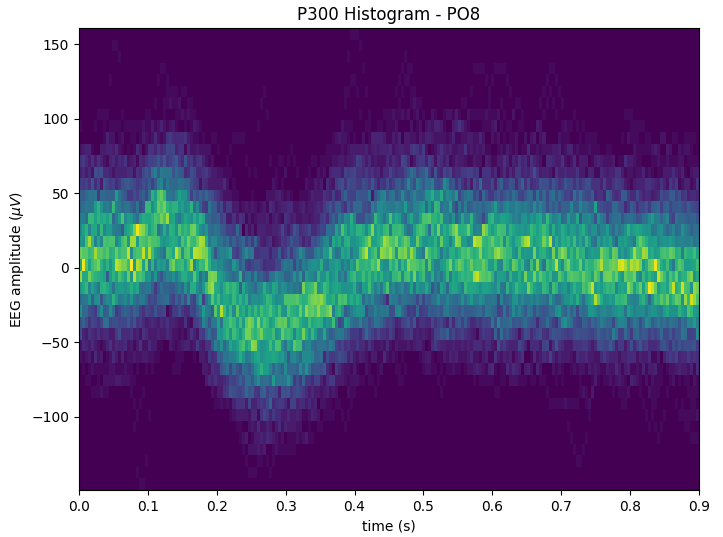}
	\caption{Histogram of a P300 wave in channel PO8 across trials of a subject, estimated with \cite{pyriemann}.}
	\label{fig:p300_wave}
\end{figure}

To overcome this limitation, we introduce a novel
B\underline{a}ye\underline{s}ian \underline{a}ccumulation of Riemannian \underline{p}robabilities (\OurModel).
This is an end-to-end pipeline for P300 BCI classification, following these successive steps:
(1) feature extraction, estimating prototype-based covariance matrices from EEG epochs,
(2) ERP-level classification, performed with a probabilistic Riemannian MDM,
(3) character-level classification, carried out with a Bayesian inference cumulating confidence after each trial.
The \OurModel{} pipeline provides an anytime computation framework,
always yielding the best results in the sense of the maximum likelihood.
One major contribution of our approach is to use both target and non-target flashes to softly update the character classification,
while existing pipelines concentrate only on target flashes.
By avoiding this loss of information, each flash contributes to updating the confidence estimation of the characters,
leading to a faster P300 BCI.


In this article, Section~\ref{sec:sota} reviews the state-of-the-art P300 BCI~\cite{barachant2014plug} and its current limitations;
Section~\ref{sec:end2end} shows the derivation of the new end-to-end pipeline.
The new pipeline is applied on real P300-speller data, as described in Section~\ref{sec:appl_p300};
and comparative results with the state-of-the-art are given in Section~\ref{sec:res}, followed by a Discussion and Conclusion.



\section{Description of P300 BCI}
\label{sec:bci}

BCI rely either on the decoding of different mental tasks or on the reaction to an external stimulation.
The former are called independent BCI and the latter dependent BCI~\cite{WOL02}.
In dependent BCI, ERP are good candidates to design synchronous BCI systems.
While error-related potentials~\cite{chavarriaga_learning_2010}, N170 (movement or face detection) or N400 (face recognition) have been investigated, the most common ERP-based BCI rely on the P300 wave~\cite{Farwell1988}.
The P300 wave is an automatic response to an oddball stimulus, \textit{i.e.}, a stimulus which is both awaited and infrequent.
The P300 wave is the result of a combination of components~\cite{luck2014introduction},
the most prominent being a peak occurring 300 ms after the stimulus, as shown in Figure~\ref{fig:p300_wave}.

\begin{figure}[t]
	\centering
	\pgfimage[width=0.8\linewidth]{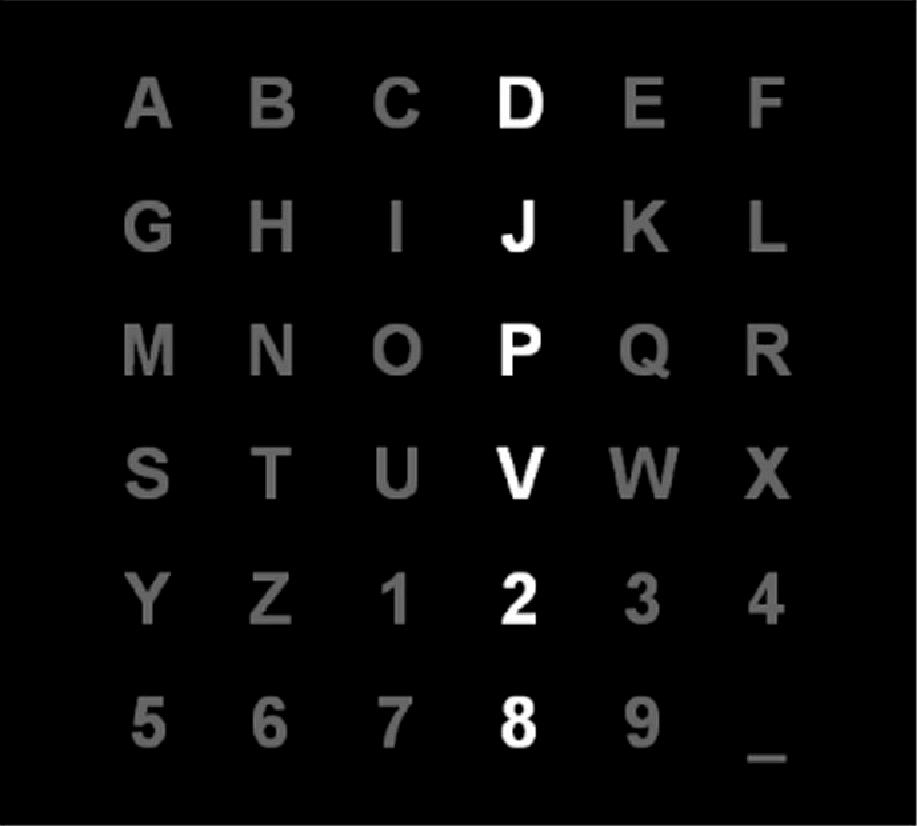}
	\caption{\hl{Virtual keyboard of a P300-speller using pseudo-random character flashes.}}
	\label{fig:p300_speller}
\end{figure}

The most common applications of P300-based BCI are spellers~\cite{Farwell1988,Rivet2009,barachant2014plug}
and games~\cite{barachant2014plug,Goncharenko2020Raccoons}.
Such systems display characters (also called symbols or items) that are flashed with a fixed interstimulus interval (ISI).
To keep all characters within the visual field of the subject, they are displayed on a grid.
Different flashing strategies have been proposed, using either row-column flashing or pseudo-random flashing.
\hl{The flashing characters are usually presented briefly in inverted colors as shown on Figure~\ref{fig:p300_speller},
but it is possible to use different stimulus~\cite{jin2012changing}.}

A trial is an EEG epoch of fixed length (often 1 s) starting at the time of the flash.
When all the characters have been flashed once, the associated group of trials is called a repetition.
P300-based BCI discriminate $K=2$ ERP classes:
target trial (denoted by $k=\text{T}$) where the chosen character has been flashed, evoking a P300 wave;
and non-target trial (denoted $k=\text{NT}$) where the chosen character has not been flashed, resulting in an absence of a P300 wave.
It is a difficult task to detect a P300 at a trial level (called single-trial detection),
this is why the character classification is done after several repetitions to accumulate evidence and increase the accuracy.
After several repetitions, the classifier makes a prediction to select a character and is ready to start over for another character prediction.

For a more complete description of P300-based BCI, the reader can refer to~\cite{Rivet2009}.

\section{State of the art for P300 BCI classification}
\label{sec:sota}

This section describes the state of the art in P300-based BCI,
combining single-trial ERP classification achieved with Riemannian MDM~\cite{barachant2014plug}
and counting occurrences of detected characters to make a prediction~\cite{Farwell1988}.
This approach yields good accuracy with short calibration time~\cite{Lotte2018}.

\subsection{Riemannian MDM for ERP classification}
\label{ssec:sota_erp_classif}

Let denote by $X \in \mathbb{R}^{C \times N}$ a trial of EEG signals, recorded on $C$ channels (or electrodes) and on $N$ temporal samples.
This EEG signal has been previously band-pass filtered between 1 and 20 Hz.
Riemannian approaches are not applied on the signal itself, but on covariance matrices estimated from this signal.
Since the discriminant information is given by the temporal waveform of P300 rather than its spatial covariance,
an extended trial $\dot{X} \in \mathbb{R}^{2C \times N}$ is built as~\cite{barachant2014plug}:
\begin{equation}
	\label{eq:ext_epoch}
	\dot{X} = \begin{bmatrix} P \\ X \end{bmatrix} \ ,
\end{equation}
where $P \in \mathbb{R}^{C \times N}$ denotes the prototype response, estimated as the grand average of target response.
Then, the covariance matrix $\Sigma \in \mathbb{R}^{2C \times 2C}$ can be estimated as:
\begin{equation}
	\label{eq:cov}
	\Sigma = \frac{1}{N-1} \dot{X} \dot{X}^T \ .
\end{equation}
Covariance matrices are symmetric positive definite (SPD), \textit{i.e.},
they have strictly positive eigenvalues and are confined in a subspace of the Euclidean space \cite{congedo2017riemannian}.
Endowed with an appropriate distance, it is possible to define a dedicated geometry to consider those matrices,
called a Riemannian manifold $\Ma$~\cite{Congedo2013HDR,Yger2017}.
An appropriate distance could be the affine-invariant,
defined between matrices $\Sigma_1$ and $\Sigma_2$ as~\cite{moakher2005differential}:
\begin{equation}
	d(\Sigma_1,\Sigma_2) = \left( \sum_{c=1}^{2C} \log^{2} e_{c} \right)^{\frac{1}{2}} \ , \nonumber
\end{equation}
where $e_c$, $c = 1, \dots, 2C$, are the eigenvalues of $\Sigma_1^{-1} \Sigma_2$.

The Riemannian minimum-distance-to-mean (MDM) is a deterministic classifier introduced in \cite{Barachant2012}.
It is a generative classifier, requiring simply to estimate one mean $\bar{\Sigma}_k$ for each class $k = 1, \dots, K$
during the training step:
\begin{equation}
	\label{eq:mdm_train}
	\bar{\Sigma}_k = \arg \min_{\Sigma \in \Ma} \sum_{i \in \mathcal{I}(k)} d^2(\Sigma_i,\Sigma) \ , \nonumber
\end{equation}
where $\mathcal{I}(k)$ is the set of indices of training matrices belonging to class $k$.
This geometric mean has no closed-form solution and therefore has to be computed iteratively \cite{fletcher2004principal}.
Thus, for a single-trial classification, the trial covariance $\Sigma$ is assigned to the class with the closest mean:
\begin{equation}
	\label{eq:mdm_class}
	\hat{k} = \arg \min_{k=1 \ldots K} \ d(\Sigma, \bar{\Sigma}_k) \ .
\end{equation}

\subsection{Maximization of occurrences for character classification}
\label{ssec:sota_character_classif}

The historical character classification is a simple decision rule based on the maximization of occurrences of detected characters,
\textit{i.e.} characters flashed when a P300 has been detected in the trial.
The character classification is performed after $t$ trials \cite{Farwell1988}:
\begin{equation}
	\label{eq:multi_p300}
	\hat{l}_t = \arg \max_{l=1 \ldots L} \ \sum_{\tau=1}^{t} \delta_{\hat{k}_{l,\tau}}^\kt \ ,
 \end{equation}
where $\delta_{a}^b$ is the Kronecker delta which equals 1 when $a = b$, and $0$ otherwise;
and where $\hat{k}_{l,\tau}$ denotes the output class/label of a single-trial ERP classification when character $l$ has been flashed during trial $\tau$.
In the case of a Riemannian MDM, ERP classification is given by Eq.~\eqref{eq:mdm_class}.

Note that the classification score in the right side of Eq.~\eqref{eq:multi_p300}
can be easily transformed into a character probability in $[0,1]$,
dividing the sum by the number of trials.

\subsection{Limitations}
\label{ssec:sota_lim}

\begin{figure}[t]
	\begin{center}
		\resizebox{11cm}{!}{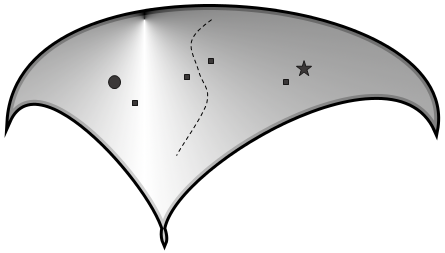}
	\end{center}
	\caption{Illustration of the classification of four trials in the space of SPD matrices,
	for two classes $\knt$ and $\kt$ with their centers $\bar{\Sigma}_\knt$ and $\bar{\Sigma}_\kt$,
	their equidistance represented by the dotted line.
	In this example, the target character has been flashed at trial $4$, and not flashed before.
	At ERP classification,
	deterministic MDM sees no difference between trials $\Sigma_3$ and $\Sigma_4$ (same class),
	whereas probabilistic MDM does (same class but different probabilities);
	deterministic MDM sees a huge difference between trials $\Sigma_3$ and $\Sigma_1$ (different classes),
	whereas probabilistic MDM understands their similarity (different classes but close probabilities).
	At character classification, the argmax-argmin classification updates its sum only during flashed trial $4$,
	whereas Bayesian accumulation updates its confidence after each trial.
	}
	\label{fig:determ_vs_proba}
\end{figure}

On the one hand, combining Eq.~\eqref{eq:mdm_class} and~\eqref{eq:multi_p300} gives an argmax-argmin character classification.
This approach is not optimal, as it loses information:
Riemannian distance to ERP-class centers yields a crucial information to quantify the confidence into the ERP classification.
This information is lost in the deterministic classification shown in Eq.~\eqref{eq:mdm_class}.
The confidence in ERP class could be transformed in a continuous probability,
allowing to softly consolidate character classification confidence along trials.
This is not possible in the argmax-argmin approach, where single-trial ERP classification error strongly impacts the character classification.
The need for a probabilistic MDM rather than a deterministic one is illustrated on Figure~\ref{fig:determ_vs_proba}.

On the other hand, for usual character classifier, the confidence in a character is increased
only when the character is flashed and a P300 is detected in the trial.
In the case of a Riemannian MDM used for ERP classification~\cite{barachant2014plug},
this occurs when the trial covariance matrix is close to the mean of the target class.
Indeed, it seems possible to increase the confidence for the target character when
the character is not flashed and the trial covariance matrix is close to the mean of non-target class.
This information is currently not captured by Eq.~\eqref{eq:multi_p300},
resulting in a loss of discriminative power since non-target flashes are more frequent than target ones.


\section{End-to-end P300 BCI classification}
\label{sec:end2end}

This section expounds the derivation of an end-to-end P300 BCI pipeline,
building on Bayesian accumulation to perform character classification.
The details of this derivation are indicated below,
providing all underlying assumptions required to obtain the final expression.

\subsection{Bayesian accumulation of character probabilities}
\label{ssec:end2end_bayesian}


In an online setup, after accumulating $\tau=1, \ldots, t$ trials of the same target character,
the cumulative posterior probability of character $l$ can be written as:
\begin{align}
	\label{eq:ct_proba}
	p(l & | X_1, \ldots, X_t) = \nonumber
	\\
	& \frac{ p(X_t | l, X_1, \ldots, X_{t-1}) \ p(l | X_1, \ldots, X_{t-1}) }{ p(X_t | X_1, \ldots, X_{t-1}) } \ ,
\end{align}
combining Bayes' rule with the chain rule of probability.
%
Since this expression is intractable, we make an approximation assuming that $X_t$ is independent from $X_1, \ldots, X_{t-1}$.
While this is true for $X_1$, it is only partially true for $X_{t-1}$, as it depends on the overlap between epochs $X_{t-1}$ and $X_t$).
Eq.~\eqref{eq:ct_proba} is simplified into:
\begin{equation}
	\label{eq:ct_prob_2}
	p(l | X_1, \ldots, X_t) = \frac{ p(X_t | l) \ p(l | X_1, \ldots, X_{t-1}) }{ p(X_t) } \ ,
\end{equation}
and, if $p(l | X_1, \ldots, X_t)$ is renamed $p_t(l)$, it becomes:
\begin{equation}
	\label{eq:ct_prob_3}
	p_t(l) = \frac{ p(X_t | l) \ p_{t-1}(l) }{ p(X_t) } \ ,
\end{equation}
where $p_{t-1}(l)$ can be seen as the prior from previous trials. 
It is dynamically updated, as the previous posterior becomes the following prior.
%
Using the law of total probability, the denominator is expanded as:
\begin{equation}
	\label{eq:ct_prob_4}
	p_t(l) = \frac{ p(X_t | l) \ p_{t-1}(l) }{ \sum_{\lambda=1}^{L} p(X_t|\lambda) \ p_{t-1}(\lambda) } \ .
\end{equation}
Since $p_{t-1}(l)$ is a fraction depending of $p_{t-2}(l)$, and $p_{t-2}(l)$ a fraction depending of $p_{t-3}(l)$, etc.,
this formula can be expanded in a finite continued fraction until $p_{0}(l)$,
which can be finally simplified into:
\begin{equation}
	\label{eq:ct_char_proba}
	p_t(l) =
	\frac{ \prod_{\tau=1}^{t} \ p(X_\tau | l) \ p_0(l) }{ \sum_{\lambda=1}^{L} \prod_{\tau=1}^{t} \ p(X_\tau | \lambda) \ p_0(\lambda) } \ .
\end{equation}
where $p_0(l)$ is the initial prior on character $l$.

\subsection{Bayesian accumulation of ERP probabilities}

Reverting the marginalization of the ERP class $k$ out of the likelihood, we have:
\begin{align}
	p(X_\tau | l) &= \sum_{k=1}^{K} \ p(X_\tau , k | l) \nonumber
	\\
	&= \sum_{k=1}^{K} p(X_\tau | k, l) \ p(k | l) \nonumber
	\\
	\label{eq:ct_char_lik}
	&= \sum_{k=1}^{K} p(X_\tau | k) \ p(k | l) \ ,
\end{align}
because epoch $X_\tau$ depends only on ERP class $k$, not on character $l$.
Considering a P300 BCI with $K=2$ ERP classes, $k=\text{T}$ for target and $k=\text{NT}$ for non-target,
Eq.~\eqref{eq:ct_char_lik} becomes:
\begin{equation}
	\label{eq:ct_p300_lik}
	p(X_\tau | l) = p(X_\tau | \kt) \ p(\kt | l) + p(X_\tau | \knt) \ p(\knt | l) \ ,
\end{equation}
where $p(\kt | l) = 1$ (\textit{resp.} $0$) and $p(\knt | l) = 0$ (\textit{resp.} $1$)
when the character $l$ has been flashed (\textit{resp.} not flashed) at trial $\tau$.
This deterministic binarization justifies the equivalence between ``target character flashed'' and ``presence of ERP in trial''.

Finally, Eq.~\eqref{eq:ct_char_proba} becomes:
\begin{align}
	\label{eq:ct_char_proba2}
	& p_t(l) =
	\\
	& \frac{ \prod_{\tau=1}^{t} \ \left( p(X_\tau | \kt) \ p(\kt | l) + p(X_\tau | \knt) \ p(\knt | l) \right) \ p_0(l) }{ \sum_{\lambda=1}^{L} \prod_{\tau=1}^{t} \ \left( p(X_\tau | \kt) \ p(\kt | \lambda) + p(X_\tau | \knt) \ p(\knt | \lambda) \right) \ p_0(\lambda) } \ . \nonumber
\end{align}
This formula allows a seamless classification for P300 BCI from EEG epochs to characters,
without requiring an explicit ERP-level classification.
It is valid for any type of probabilistic and generative ERP classifier returning $p(X_\tau | k)$,
and thus could be used tomorrow with deep neural networks \cite{Roy2019}.

\begin{table*}[t]
  \centering
  \begin{tabular}{|l|c|c|c|c|c|}
    \hline
    Name & \# Channels & \# Subjects & \# Characters per subject & \# Trials per character (T / NT) & Reference
		\rule[-7pt]{0pt}{20pt} \\ \hline \hline
    BNCI 2014-008 & 8 & 8 & 35 & 20 / 100 & \cite{riccio2013attention}
		\rule[-7pt]{0pt}{20pt} \\ \hline
    BNCI 2014-009 & 16 & 10 & 18 & 16 / 80 & \cite{aloise_covert_2012}
		\rule[-7pt]{0pt}{20pt} \\ \hline
  \end{tabular}
  \caption{Details of datasets used for evaluation of P300-speller.}
  \label{tab:datasets}
\end{table*}

\subsection{Bayesian accumulation of Riemannian probabilities}
\label{ssec:end2end_riemannian}

Since Riemannian classifiers obtain best results for ERP classification~\cite{barachant2014plug,Lotte2018},
we derive a probabilistic version of Riemannian MDM, called pMDM.
Assuming that epoch $X$ follows a multivariate normal model $\mathcal{N}(0,\Sigma)$,
with zero mean (EEG signals are centered after a band-pass filtering) and with a covariance matrix $\Sigma \in \Ma$,
we define $p(X | k) = p(\Sigma | k)$.

To represent covariance matrices belonging to a Riemannian space, the Riemannian Gaussian distribution can be used to model each class.
Its probability density function is defined as~\cite{Said2017}:
\begin{equation}
	\label{eq:riem_gauss}
	p(\Sigma | k) = p(\Sigma | \bar{\Sigma}_k,\sigma_k) = \frac{1}{\zeta(\sigma_k)} \exp \left(- \frac{d^2(\Sigma,\bar{\Sigma}_k)}{2 \sigma^2_k} \right) \ ,
\end{equation}
where $\bar{\Sigma}_k \in \Ma$ is the center of the Riemannian Gaussian distribution of class $k$,
$\sigma_k>0$ encodes the dispersion of the distribution, and $\zeta(\sigma_k)$ is a normalization factor.
Under the assumption that all classes have identical dispersion $\sigma_k$, the Riemannian instantiation of Eq.~\eqref{eq:ct_char_proba2} is:
\begin{equation}
	\label{eq:ct_p300_proba_2}
	p_t(l) =
	\frac{ \prod_{\tau=1}^{t} \ e^{- d^2(\Sigma_\tau, \tilde{\Sigma}_{l,\tau})} \ p_0(l) }{ \sum_{\lambda=1}^{L} \prod_{\tau=1}^{t} \ e^{- d^2(\Sigma_\tau, \tilde{\Sigma}_{\lambda,\tau})} \ p_0(\lambda) } \ ,
\end{equation}
where $\tilde{\Sigma}_{l,\tau} = \bar{\Sigma}_{\kt}$ if character $l$ has been flashed during trial $\tau$, and
$\tilde{\Sigma}_{l,\tau} = \bar{\Sigma}_{\knt}$ otherwise.

After trial $t$, the character classification is obtained thanks to the maximum \textit{a posteriori} (MAP),
\textit{i.e.} maximizing the posterior probability $p(l | X_1, \ldots, X_t)$.
The MAP classification rule is:
\begin{align}
	\label{eq:ct_p300_map}
	\hat{l}_t & = \arg \max_{l=1 \ldots L} \ p_t(l) \nonumber
	\\
	& = \arg \max_{l=1 \ldots L} \ \prod_{\tau=1}^{t} \ e^{- d^2(\Sigma_\tau, \tilde{\Sigma}_{l,\tau})} \ p_0(l) \ ,
\end{align}
because the denominator is independent from $l$.
Priors $p_0(l)$ can easily be used to embed letter and word prediction in BCI \cite{Kaufmann2012}.
Under the assumption of equally likely classes, \textit{i.e.} equal priors $p_0(l)$, it becomes a maximum likelihood (ML) classification rule:
\begin{equation}
	\label{eq:ct_p300_ml}
	\hat{l}_t = \arg \min_{l=1 \ldots L} \ \sum_{\tau=1}^{t} \ d^2(\Sigma_\tau, \tilde{\Sigma}_{l,\tau}) \ .
\end{equation}

Contrary to the argmax-argmin approach, this end-to-end derivation leads to a single argmin approach for the ML rule,
which is a seamless processing of information from EEG epochs to BCI characters.
It prevents loss of information because all characters softly update their confidence on both target and non-target flashes,
contributing to increasing the confidence of the target character.

\begin{figure*}[t!]
	\begin{center}
		\pgfimage[width=0.95\linewidth]{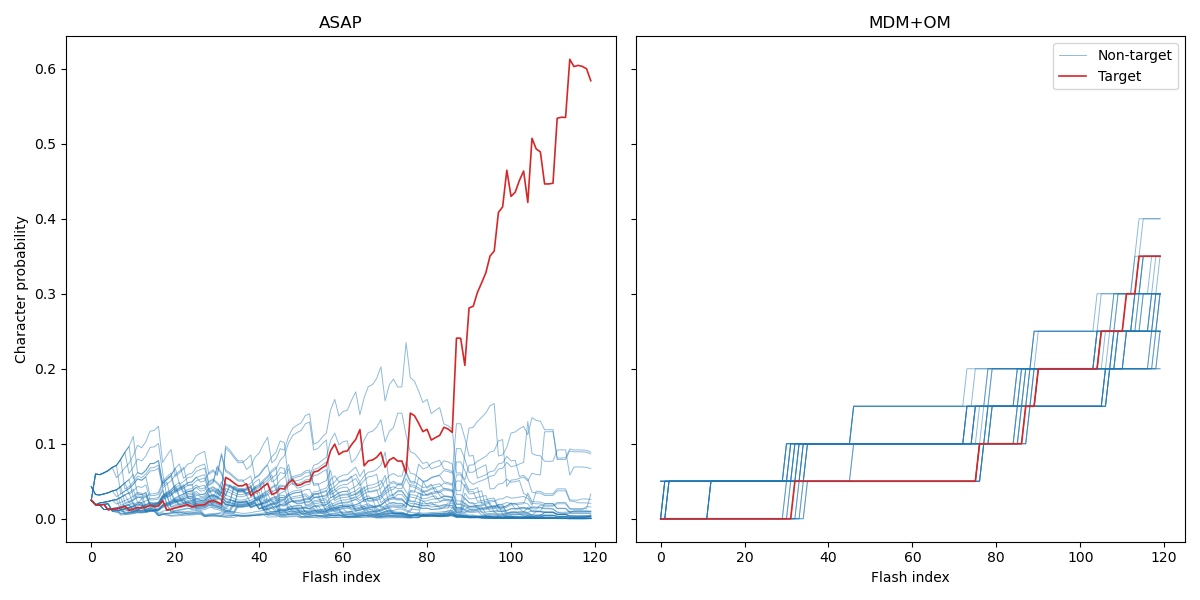}
	\end{center}
	\caption{Character probabilities $p_t(l)$ as a function of flash $t$, for only one character classification,
	for the target character in red and the non-target characters in blue.
	Left: \OurModel{}, softly updated after each flash. Right: \MdmModel{}, hardly updated only on target flashes.
	}
	\label{fig:proba}
\end{figure*}

\begin{figure*}[tbh]
	\begin{center}
		\pgfimage[width=0.95\linewidth]{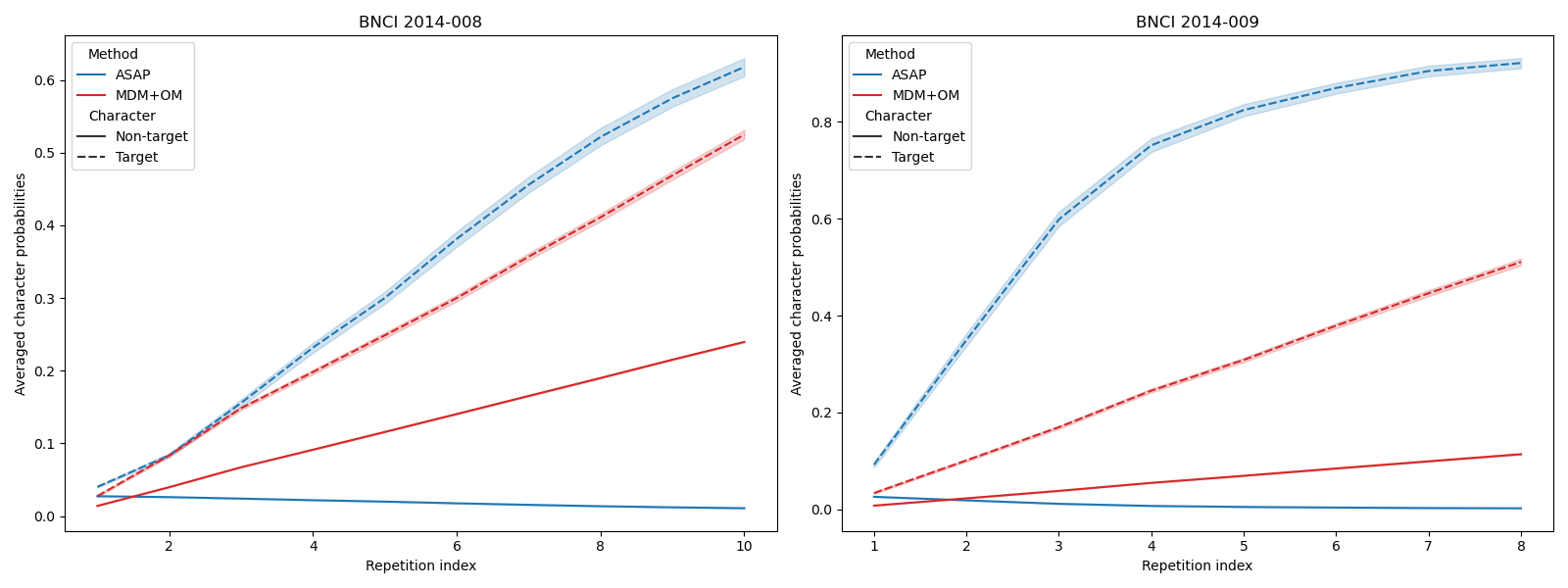}
	\end{center}
	\caption{Character probabilities as a function of repetition,
	for \MdmModel{} and \OurModel{}, averaged across all character classifications of each dataset.
	}
	\label{fig:proba_average}
\end{figure*}


\subsection{Contributions summary}

This B\underline{a}ye\underline{s}ian \underline{a}ccumulation of Riemannian \underline{p}robabilities is called \OurModel.
This online algorithm must be re-initialized at the beginning of each new character classification,
and it can be stopped with any dynamic stopping strategy \cite{Schreuder2013,Kindermans2014}.

\OurModel{} is an end-to-end P300 BCI, with a classification scheme occurring at character-level,
which is the actual end of ERP-based BCI.
Most of the existing approaches are assessed at ERP-level,
\textit{e.g.} as detailed in~\cite{barachant2014plug,korczowski2015single,congedo2017riemannian}.
While these are useful for signal processing and classification,
they do not encompass the character classification which was regarded as an engineering issue.
We demonstrate in this article that a principled and theoretical approach is possible,
and that combining Riemannian geometry and Bayesian inference allow us to devise a character classification pipeline,
integrating ERP probabilization (instead of binarization) in the process.

Note that taking into account non-target flashes has already been considered for P300 classification
with a Bayesian framework~\cite{Dauce2014}, but for a discriminative classifier.
This kind of classifier provides $p(k|X_t)$ instead of $p(X_t|k)$ for generative ones~\cite{Ng2002},
and requires a large training set to avoid overfitting, \emph{e.g.} 264,000 trials are used in~\cite{Dauce2014}),
which prevents reproducibility, scalability and out-of-the-lab deployment.

\begin{figure*}[tbh]
	\begin{center}
		\pgfimage[width=0.95\linewidth]{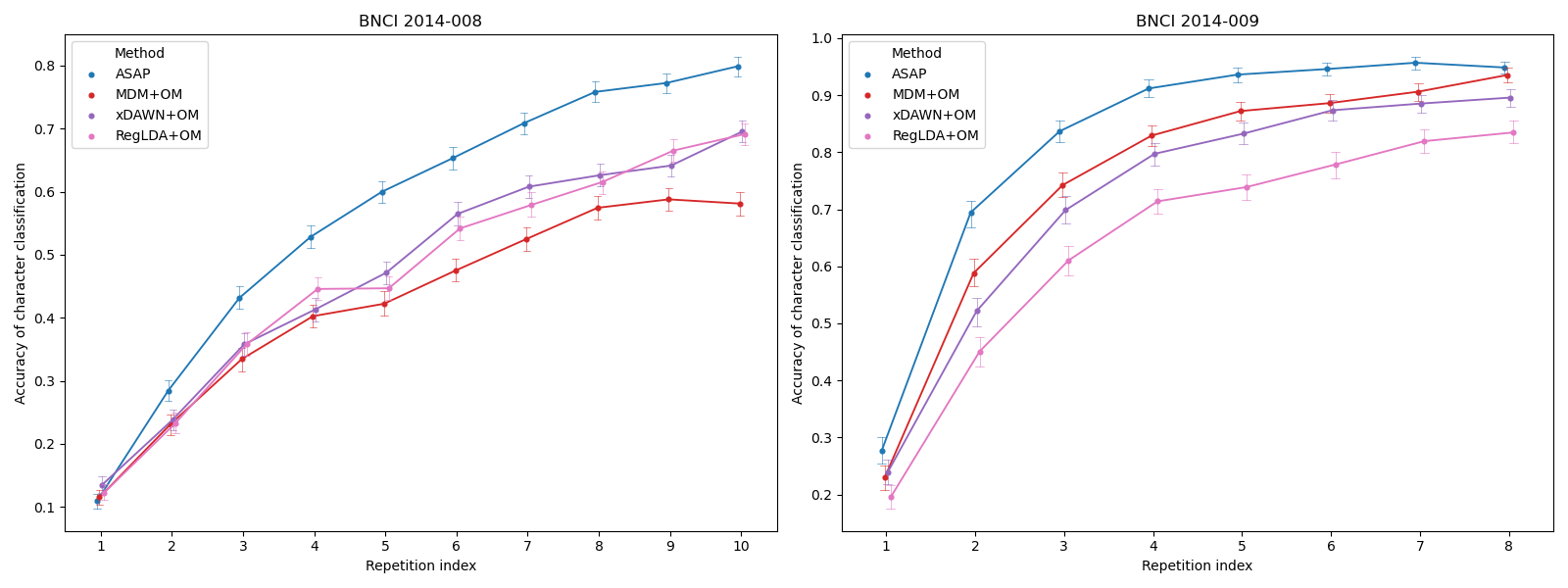}
	\end{center}
	\caption{\hl{Character classification accuracy as a function of repetition, for \OurModel{}, \MdmModel{}, \XdawnModel{} and \LdaModel{},
	averaged across all character classifications of both datasets.}
	}
	\label{fig:accuracy}
\end{figure*}

\section{Application to P300-speller}
\label{sec:appl_p300}

The introduced pipeline can be applied to all ERP-based BCI.
As validation, it is applied here on the P300-speller task described in Section~\ref{ssec:sota_character_classif}.
While most works evaluate their performances at ERP-level \cite{barachant2014plug,korczowski2015single,congedo2017riemannian},
we evaluate them at character-level, which is the actual end of the BCI pipeline.

\subsection{Data}

EEG data come from 2 datasets containing 18 subjects and gathering 50,880 trials. 
The important information of the datasets is provided in Table~\ref{tab:datasets}.

\subsubsection{BNCI 2014-008 dataset}

This EEG dataset is publicly available~\cite{riccio2013attention} and gathers the recordings of 8 patients
with amyotrophic lateral sclerosis visually focusing on a $6 \time 6$-letter matrix.
The data are recorded with a $C=8$ channels amplifier (g.MOBILAB, g.tec, Austria)
using active electrodes (g.Ladybird, g.tec, Austria)
that are located at Fz, Cz, Pz, Oz, P3, P4, PO7 and PO8.
Reference is the right ear lobe and the ground is taken from the left mastoid.
The signal is acquired at 256 Hz.
The speller randomly highlights the lines and columns from the letter matrix for 125 ms
followed by 125 ms of ISI, resulting in a stimulus onset asynchrony of 250 ms.

\subsubsection{BNCI 2014-009 dataset}

This EEG dataset is publicly available~\cite{aloise_covert_2012} and gathers the recordings of 10 healthy subjects
visually focusing on a $6 \time 6$-letter matrix.
The data are recorded with a $C=16$ electrodes that are located at
Fz, FCz, Cz, CPz, Pz, Oz, F3, F4, C3, C4, CP3, CP4, P3, P4, PO7, and PO8.
Each electrode is referenced to the linked earlobes and the ground is taken from the right mastoid.
The EEG was acquired at 256 Hz.



\subsection{Methods}

\hl{Raw EEG signals are band-pass filtered between 1 to 20 Hz.
Trials are extracted as EEG epochs of fixed length (1 s) starting at the time of flashes.}

\hl{Four single-trial character-level classifiers for P300 BCI are compared in this article:}
\begin{itemize}
\item \hl{ASAP with equal priors, \textit{i.e.} the ML version in Eq.~\eqref{eq:ct_p300_ml},
which is the novel Bayesian accumulation (BA) of probabilistic Riemannian MDM, a kind of pMDM+BA pipeline;}
\item \hl{\MdmModel{}: Riemannian MDM \cite{barachant2014plug}, see Eq.~\eqref{eq:mdm_class},
followed by occurrences maximizing (OM), see Eq.~\eqref{eq:multi_p300},
which is the state of the art described in Section~\ref{sec:sota};}
\item \hl{\XdawnModel{}: downsampling at 128 Hz, followed by xDAWN \cite{Rivet2009}, regularized LDA on vectorized data, and OM;}
\item \hl{\LdaModel{}: downsampling at 128 Hz, followed by regularized LDA on vectorized data \cite{Krusienski2006,Blankertz2011}, and OM.}
\end{itemize}

\subsection{Evaluation}

The current experiment is a within-session classification of P300-speller characters,
comparing evolution of performances (probabilities, accuracy and BCI ITR \cite{Yuan2013})
along time (flash and repetition).
For each session of each subject,
the training set contains the trials associated to the first 6 characters
(to calibrate the $K=2$ class centers $\bar{\Sigma}_{\kt}$ and $\bar{\Sigma}_{\knt}$ for \MdmModel{} and \OurModel{}),
and the test set contains remaining trials.

The code of the experiment relies on the rich ecosystem of open source libraries available for
BCI datasets, EEG signal processing and machine learning, that is
Timeflux~\cite{timeflux_2019}, MOABB~\cite{Jayaram2018}, MNE~\cite{gramfort_mne_2014},
scikit-learn~\cite{scikit-learn} and pyRiemann~\cite{pyriemann}.

\hl{For research reproducibility, code of offline experiments is available at \url{https://github.com/sylvchev/asap-p300-bci}.
Open-source implementation of \OurModel{} is available as an online Timeflux \cite{timeflux_2019} application:
\url{https://github.com/timeflux/demos/tree/main/speller/P300}.
This free and open framework allows to evaluate the proposed implementation in a real online setup.}

%


\section{Results}
\label{sec:res}

Figure~\ref{fig:proba} displays the character probabilities $p_t(l)$ as a function of flash $t$,
for only one character classification.
The target character is in red and non-target characters are in blue.
On the right side, we observe that, for \MdmModel{}, only probabilities of flashed characters are updated,
and that probabilities of non-target characters increase with time.
In this example, the classification will be incorrect because target character does not have the highest probability.
On the left part, we can see that all probabilities of \OurModel{} are updated after each flash:
it softly consolidates confidence across target flashes, but also non-target ones, which are much more frequent.
We also see that probabilities of non-target characters decrease over time
and actually contribute to increasing confidence in the target character.
Dependence between character probabilities comes the denominator of Eq.~\eqref{eq:ct_p300_proba_2}.

Character probabilities are averaged across all character classifications of each dataset.
Figure~\ref{fig:proba_average} shows the character probabilities as a function of repetition,
with the average and its confidence interval at 95 \%.
For \MdmModel{}, the averaging has smoothed the steps visible on the right part of Figure~\ref{fig:proba}.
This figure confirms that \OurModel{} increases probabilities of target characters faster than \MdmModel{},
and that \OurModel{} decreases probabilities of non-target characters contrary to \MdmModel{}.
Combining these two properties provides a faster and larger divergence between probabilities of target and non-target characters.

Figure~\ref{fig:accuracy} displays the accuracy of character classification performed after each repetition,
averaged across all character classifications of each dataset.
We can see that trends observed in probabilities have a strong impact on classification accuracy.
Divergences between probabilities allow for a better discrimination between target and non-target characters.
The first dataset (left side) is more challenging than the second one (right side) where subjects had previous experience with P300-based BCI,
\hl{but \OurModel{} is always better and faster than \MdmModel{}, \XdawnModel{} and \LdaModel{}.
For a fixed number of repetitions, \OurModel{} gives a better classification than all other methods;
and for a fixed level of confidence \OurModel{} allows for a faster classification than all other methods.}
%

\begin{figure*}[t]
	\begin{center}
		\pgfimage[width=0.95\linewidth]{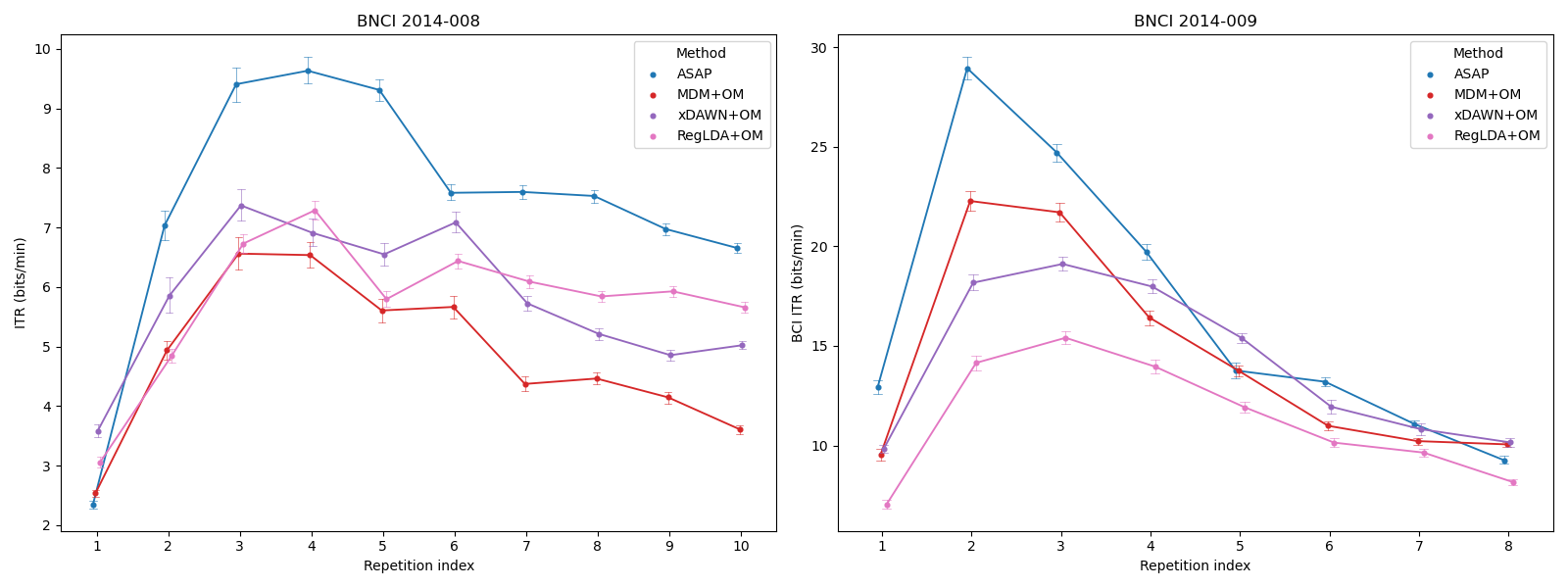}
	\end{center}
	\caption{\hl{BCI ITR (in bits/min) as a function of repetition, for \OurModel{}, \MdmModel{}, \XdawnModel{} and \LdaModel{},
	averaged across all character classifications of each dataset.}
	}
	\label{fig:itr}
\end{figure*}

%
Figure~\ref{fig:itr} displays the BCI ITR in bits/min computed after each repetition,
averaged across all character classifications of each dataset.
For each dataset, methods reach their optimal peak of BCI ITR for the same number of repetitions.
\hl{We can see that differences previously observed in accuracy have an obvious impact on BCI ITR:
\OurModel{} is always higher than \MdmModel{}, \XdawnModel{} and \LdaModel{}, giving better performance of communication.}

\hl{One the one hand, the \LdaModel{} and \XdawnModel{} methods use the same classifier, a regularized LDA.
It is applied on downsampled trials for \LdaModel{}, and on downsampled and spatially filtered trials for \XdawnModel{}.
Results show that xDAWN spatial filter, applying a supervised dimension reduction, enhances most of the time classification performances.}
On the other hand, \MdmModel{} and \OurModel{} methods use
the same features (covariance matrices $\Sigma$),
the same training models (covariance centers $\bar{\Sigma}_{\kt}$ and $\bar{\Sigma}_{\knt}$),
and the same dissimilarity measure between features (Riemannian distance).
Thus, all observed differences come only from the way to exploit information,
from the distance between covariance matrices to character classes.
While \MdmModel{} suffers of all limitations described in Section~\ref{ssec:sota_lim}, like the argmax-argmin character classification,
\OurModel{} consolidates character confidences using probabilities which are softly accumulated after each trial,
that is illustrated on the left side of Figure~\ref{fig:proba}.

\hl{To conclude these experiments performed on real data,
we have shown that \OurModel{} outperforms \MdmModel{}, \XdawnModel{} and \LdaModel{}, providing a better and faster P300 BCI.}
It reduces the number of repetitions necessary for the correct classification of characters.
\hl{Numerical values of figures can be found in Appendix.}

\section{Discussion}
\label{sec:dis}


By updating confidence on non-target flashes too,
this approach aims to exploit all information available in EEG trials.
There are several advantages, as it reduces the time required to spell a letter,
it preserves the concentration of the subject by avoiding the tiredness of seeing multiple flashes,
and it avoids wasting data, as the storage of EEG can be optimized.
%
In experiments, we have compared evolution of performances along time.
We have not used dynamic stopping ~\cite{Schreuder2013,Kindermans2014},
because compared methods are independent of it.
Adding this step will be done in a future work.

\hl{Note that Eq.~\eqref{eq:ct_char_proba} has been derived for any type of online classification.
So it could be applied to other BCI paradigms, especially asynchronous BCI making the assumption
of temporal persistence in the class across several epochs,
like asynchronous SSVEP \cite{Kalunga2016}, motor imagery \cite{Barachant2013} or neurofeedback \cite{Arns2017}.
More generally, the Bayesian scheme derived in Section~\ref{ssec:end2end_bayesian} makes no assumption
on the type of input data $X$ and is independent from the two-level classification specific to P300 BCI.
Consequently, it remains valid for any type of sequential data.
Derived in this article for multivariate biological time series, it could be applied to other data
such as audio stream \cite{Sen2019},
video stream \cite{Wang2002}, 
\textit{etc}.

Underlying assumptions of the derivation of Section~\ref{sec:end2end} have been explicitly listed, in order to be reconsidered in future works.
Notably, the deterministic binarization performed after Eq.~\eqref{eq:ct_p300_lik} does not take into account that
a lack of concentration of the subject can prevent ERP to be elicited, damaging BCI performances.
A new generation of algorithms should be able to break this simple deterministic link between flash and ERP,
and to model uncertainty thanks to continuous probabilities for $p(\kt | l)$ and $p(\knt | l)$,
which could be defined monitoring concentration level in real-time.}


\section{Conclusion}

Using a Bayesian accumulation of Riemannian probabilities, this article introduces an end-to-end P300 BCI,
providing a seamless processing of information from EEG epochs to BCI characters.
The derivation of this end-to-end pipeline details all underlying assumptions required to obtain the final formula,
for further potential generalizations.
Validated on real EEG data, this new P300 BCI is better and faster than the state-of-the-art,
reducing the number of repetitions necessary for the correct classification of characters.

Future works are to combine this new approach with the following features:
online adaptation of centers of class~\cite{barachant2014plug};
cross-subject transfer learning of centers of class~\cite{Kalunga2018,Zanini2018,Rodrigues2019};
dynamic stopping for early classification~\cite{Schreuder2013,Kindermans2014};
and letter or word prediction, embedded in BCI thanks to character-dependent priors $p_0(l)$, to boost BCI performances~\cite{Kaufmann2012}.


\section{Disclosure of interest}

The authors report there are no competing interests to declare.
This work was not supported by any grant nor funding and QB/RBL/PC have conducted this research during their spare time.


\appendices

\appendix[Numerical values]

\hl{Numerical values of Figure~\ref{fig:accuracy} are given in Table~\ref{tab:accuracy},
and Figure~\ref{fig:itr} in Table~\ref{tab:itr}.}

\begin{table*}[t]
  \centering
  \begin{tabular}{|l|c|c|c|c|c|c|c|c|c|c|}
    \hline
       Repetition index & 1 & 2 & 3 & 4 & 5 & 6 & 7 & 8 & 9 & 10
       \rule[-7pt]{0pt}{20pt} \\ \hline
       \hline
       BNCI 2014-008, \OurModel{}& 0.11 & 0.28 & 0.43 & 0.53 & 0.6 & 0.65 & 0.71 & 0.76 & 0.77 & 0.8
       \rule[-7pt]{0pt}{20pt} \\ \hline
       BNCI 2014-008, \MdmModel{}& 0.12 & 0.23 & 0.33 & 0.4 & 0.42 & 0.47 & 0.52 & 0.57 & 0.59 & 0.58
       \rule[-7pt]{0pt}{20pt} \\ \hline
       BNCI 2014-008, \XdawnModel{} & 0.13 & 0.24 & 0.36 & 0.41 & 0.47 & 0.56 & 0.61 & 0.63 & 0.64 & 0.7
       \rule[-7pt]{0pt}{20pt} \\ \hline
       BNCI 2014-008, \LdaModel{} & 0.12 & 0.23 & 0.36 & 0.45 & 0.45 & 0.54 & 0.58 & 0.62 & 0.67 & 0.69
       \rule[-7pt]{0pt}{20pt} \\ \hline
       BNCI 2014-009, \OurModel{} & 0.28 & 0.69 & 0.84 & 0.91 & 0.94 & 0.95 & 0.96 & 0.95 & \multicolumn{2}{c}{}
       \rule[-7pt]{0pt}{20pt} \\ \cline{1-9}
       BNCI 2014-009, \MdmModel{} & 0.23 & 0.59 & 0.74 & 0.83 & 0.87 & 0.89 & 0.91 & 0.94 & \multicolumn{2}{c}{}
       \rule[-7pt]{0pt}{20pt} \\ \cline{1-9}
       BNCI 2014-009, \XdawnModel{} & 0.24 & 0.52 & 0.7 & 0.8 & 0.83 & 0.87 & 0.89 & 0.9 & \multicolumn{2}{c}{}
       \rule[-7pt]{0pt}{20pt} \\ \cline{1-9}
       BNCI 2014-009, \LdaModel{} & 0.2 & 0.45 & 0.61 & 0.71 & 0.74 & 0.78 & 0.82 & 0.83 & \multicolumn{2}{c}{}
       \rule[-7pt]{0pt}{20pt} \\ \cline{1-9}
  \end{tabular}
  \caption{\hl{Numerical values of Figure~\ref{fig:accuracy}: averaged character classification accuracy as a function of repetition.}}
  \label{tab:accuracy}
\end{table*}

\begin{table*}[t]
  \centering
  \begin{tabular}{|l|c|c|c|c|c|c|c|c|c|c|}
    \hline
    Repetition index & 1 & 2 & 3 & 4 & 5 & 6 & 7 & 8 & 9 & 10
    \rule[-7pt]{0pt}{20pt} \\ \hline
    \hline
       BNCI 2014-008, \OurModel{} & 2.4 & 7.0 & 9.4 & 9.6 & 9.3 & 7.6 & 7.6 & 7.5 & 7.0 & 6.7
       \rule[-7pt]{0pt}{20pt} \\ \hline
       BNCI 2014-008, \MdmModel{} & 2.5 & 4.9 & 6.6 & 6.5 & 5.6 & 5.7 & 4.4 & 4.5 & 4.1 & 3.6
       \rule[-7pt]{0pt}{20pt} \\ \hline
       BNCI 2014-008, \XdawnModel{} & 3.6 & 5.9 & 7.4 & 6.9 & 6.5 & 7.1 & 5.7 & 5.2 & 4.9 & 5.0
       \rule[-7pt]{0pt}{20pt} \\ \hline
       BNCI 2014-008, \LdaModel{} & 3.1 & 4.8 & 6.7 & 7.3 & 5.8 & 6.4 & 6.1 & 5.8 & 5.9 & 5.7
       \rule[-7pt]{0pt}{20pt} \\ \hline
       BNCI 2014-009, \OurModel{} & 12.9 & 28.9 & 24.7 & 19.7 & 13.8 & 13.2 & 11.1 & 9.3 & \multicolumn{2}{c}{}
       \rule[-7pt]{0pt}{20pt} \\ \cline{1-9}
       BNCI 2014-009, \MdmModel{} & 9.6 & 22.3 & 21.7 & 16.4 & 13.8 & 11.0 & 10.2 & 10.1 & \multicolumn{2}{c}{}
       \rule[-7pt]{0pt}{20pt} \\ \cline{1-9}
       BNCI 2014-009, \XdawnModel{} & 9.8 & 18.2 & 19.1 & 18.0 & 15.4 & 11.9 & 10.8 & 10.2 & \multicolumn{2}{c}{}
       \rule[-7pt]{0pt}{20pt} \\ \cline{1-9}
       BNCI 2014-009, \LdaModel{} & 7.0 & 14.2 & 15.4 & 14.0 & 11.9 & 10.1 & 9.6 & 8.2 & \multicolumn{2}{c}{}
       \rule[-7pt]{0pt}{20pt} \\ \cline{1-9}
  \end{tabular}
  \caption{\hl{Numerical values of Figure~\ref{fig:itr}: averaged BCI ITR as a function of repetition.}}
  \label{tab:itr}
\end{table*}





%

\bibliographystyle{IEEEtran}
\bibliography{RiemannianGeometry}




\end{document}

%% file: Fig_Classif_DeterministicVsProbabilistic.pdf_t
\begin{picture}(0,0)%
		\includegraphics{"Fig_Classif_DeterministicVsProbabilistic"}%
\end{picture}%
\setlength{\unitlength}{4144sp}%
\begingroup\makeatletter\ifx\SetFigFont\undefined%
\gdef\SetFigFont#1#2#3#4#5{%
  \reset@font\fontsize{#1}{#2pt}%
  \fontfamily{#3}\fontseries{#4}\fontshape{#5}%
  \selectfont}%
\fi\endgroup%
\begin{picture}(6619,4185)(3016,-5551)
\put(4250,-3250){\makebox(0,0)[b]{\smash{{\SetFigFont{12}{14.4}{\familydefault}{\mddefault}{\updefault}{\color[rgb]{0,0,0}$\bar{\Sigma}_\knt$}%
}}}}
\put(6890,-3160){\makebox(0,0)[b]{\smash{{\SetFigFont{12}{14.4}{\familydefault}{\mddefault}{\updefault}{\color[rgb]{0,0,0}$\bar{\Sigma}_\kt$}%
}}}}
\put(5720,-3200){\makebox(0,0)[b]{\smash{{\SetFigFont{12}{14.4}{\familydefault}{\mddefault}{\updefault}{\color[rgb]{0,0,0}$\Sigma_3$}%
}}}}
\put(5050,-3400){\makebox(0,0)[b]{\smash{{\SetFigFont{12}{14.4}{\familydefault}{\mddefault}{\updefault}{\color[rgb]{0,0,0}$\Sigma_1$}%
}}}}
\put(6300,-3750){\makebox(0,0)[b]{\smash{{\SetFigFont{12}{14.4}{\familydefault}{\mddefault}{\updefault}{\color[rgb]{0,0,0}$\Sigma_4$}%
}}}}
\put(4800,-4000){\makebox(0,0)[b]{\smash{{\SetFigFont{12}{14.4}{\familydefault}{\mddefault}{\updefault}{\color[rgb]{0,0,0}$\Sigma_2$}%
}}}}
\end{picture}%